\documentclass{article}
\usepackage{hyperref}
\usepackage{spconf,amsmath,graphicx,graphics}
\usepackage[noblocks]{authblk}
\graphicspath{{figures/}}
\usepackage{subfigure}
\newcommand{\comment}[1]{}

\title{Comparison of Methods in Skin Pigment Decomposition}
\author[$\dag$]{\normalsize Hao GONG}
\author[$\ddag$]{\normalsize Michel DESVIGNES}
\affil[$\dag$]{{\small Arcadia, Los Angeles County, CA 91006}}
\affil[$\ddag$]{{\small GIPSA-Lab, Universit\'e Grenoble Alpes, 11 rue des Math\'ematiques, F-38400, France}}

\date{}
\begin{document}
%
\maketitle
\begin{abstract}
Decomposition of skin pigment plays an important role in medical fields. Human skin can be decomposed into two primitive components, hemoglobin and melanin. It is our goal to apply these results for diagnosis of skin cancer. In this paper, various methods for skin pigment decomposition are reviewed comparatively and the performance of each method is evaluated both theoretically and experimentally. In addition, isometric feature mapping (Isomap) is introduced in order to improve the dimensionality reduction performance in context of skin pigment decomposition.
\end{abstract}
\section{Introduction}
\label{sec:intro}
Human skin is a turbid media with multi-layered structure. Among various pigments contained in the media, hemoglobin and melanin are two primitive components\cite{r1}. A variety of familiar phenomena, such as darker after tanning, redder after alcohol assumption are always responses to the quantitative changes of hemoglobin and melanin. Also, many skin diseases (e.g., melanoma) are induced by absence or overmuch of these two pigments. This makes it necessary and valuable to estimate melanin and hemoglobin distribution objectively. Existing relative researches mainly focus on the following three aspects.

{\it Skin color analysis in color spaces}. Color is easier to perceive than reflectance spectrum. Intuitive color spaces, like HSV, allow convenient measurement of skin color. Dae Hyun Kim, etc.\cite{r2}, propose to decompose skin pigments in HSV color space from a single digital image. Though cone-shaped model and radially sorted hue values in HSV color space result in hemoglobin and melanin maps closer to the real distribution of the human subject, it is just a description rather than a true quantitative science without any relation to the optical properties of the skin.

{\it Statistical skin pigment analysis}. Principal component analysis (PCA) and independent component analysis (ICA) are applied to skin pigment decomposition in density domain by Tsumura etc.\cite{r3}. However, the direction of the color plane extracted by PCA varies as the sample skin region changes. Therefore, their approach fails in producing absolute quantity of the pigments from different samples of a human subject; rather, it produces relative quantities varying from one sample to the other.

{\it Skin chromophore quantification based on physical model}.  A Beer-Lambert
law based model-fitting technique is employed to perform more accurate skin pigment decomposition by Gong etc.\cite{r7}, where tabulated extinction coefficients of predominant chromophores is introduced to degenerate the problem into solving a system of linear equations. Although this method relies on the physical properties of skin optics and experimentally compiled absorption coefficients of skin chromophores, its limitations stem from discrepancies between the {\it in-vivo} conditions for obtaining dermoscopic images and the {\it in-situ} or {\it ex-vivo} conditions used to derive these coefficients. Additionally, it fails to consider how different light wavelengths penetrate the skin at varying depths. As a result, the conclusions reached do not apply under the conditions outlined in this paper, excluding this method from our comparative analysis and discussion.

In this paper, motivated by the methodology of Tsumura’s method\cite{r3}, we introduce Isomap\cite{r4} instead of PCA to perform the dimensionality reduction in optical density domain before ICA is applied. Isomap is based on classical multidimensional scaling (MDS) but seeks to preserve the intrinsic geometry of input data, as captured in geodesic manifold distances between all pairs of datapoints. Isomap outperforms other traditional linear techniques such as PCA on nonlinear manifold on which the input data lies. However, the application of Isomap is restricted by its high computational complexity.

\section{Skin Pigment Decomposition}
\label{sec:decomp}

\subsection{Skin Pigment Decomposition in HSV}
\label{ssec:kim}
\begin{figure}[htb]
\begin{minipage}[t]{1.0\linewidth}
  \centering
  \centerline{\includegraphics[width=1.0\linewidth]{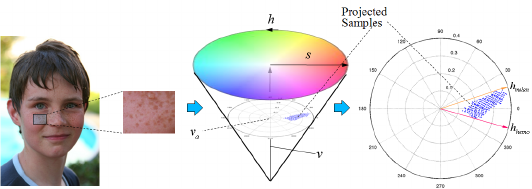}}
\end{minipage}\\
\begin{minipage}[t]{1.0\linewidth}
  \centering
  \centerline{\footnotesize (a) Sampling Region \quad\quad\quad (b) Color Model \quad\quad\quad (c) Projected Plane}
\end{minipage}
\caption{Overview of Kim’s method.}
\label{fig:kim}
\end{figure}

In this section, we briefly review the Kim’s method\cite{r2} for skin pigment decomposition in HSV color space. One visualization model of the HSV is cone. In this representation, {\it hue} is depicted as the color wheel of a 3D cone; {\it saturation} is represented by the distance from the center of a circular cross-section of the cone as shown in Fig.\ref{fig:kim}(b), and {\it value} is the distance from the apex of the cone representing the brightness. The user selected sample region $\mbox{\boldmath $R$}$, from the body reflection image, is shown in Fig.\ref{fig:kim}(a). RGB color values $(r_{i}, g_{i}, b_{i})$ of the pixels $p_{i}\in\mbox{\boldmath $R$}$ are mapped to HSV color space $(h_{i}, s_{i}, v_{i})$. By averaging $v$, the third component of HSV color space, we can find the color plane (i.e., a particular circular cross-section of the HSV color cone) $(h, s, v_{a})$, where $v_{a}=\sum_{i=1}^{n} v_{i} / n$. It can be assumed that this color plane contains the two primitive components, melanin and hemoglobin, based on the following facts: the color plane perpendicular to the value axis contains all hue and saturation informations; color planes from different human subjects or from varying sample regions are different only in their brightness. Then, the sample colors are projected along vector $\mbox{\boldmath $p$}=\left[h_{i}, s_{i}, v_{i}\right]^\top$ on to the color plane, as shown in Fig.\ref{fig:kim}. Hemoglobin and melanin components are then extracted successively by projecting the colors onto the color plane to two independent axes. Since HSV color plane, well sorts out its hue values radically around its center and sample color data is distributed nearly in the same area of red to green hues, it is intuitionistic for the user to perceive where the independent color vectors are, without counting on a statistical method such as ICA; this can be procedurally done simply by radially searching the hue values, that encompass the color samples, on the color plane projected, as shown in Fig.\ref{fig:kim}(c).

\subsection{Skin Pigment Decomposition Using PCA and ICA}
\label{ssec:tsu}
\begin{figure}[htb]
  \centering
\begin{minipage}[t]{1.0\linewidth}
  \centerline{\includegraphics[width=0.4\linewidth]{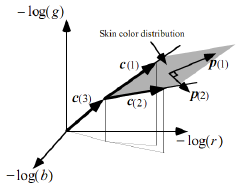}}
\end{minipage}
\caption{Skin color model in the optical density domain.}
\label{fig:optic}
\end{figure}

The Tsumura's method\cite{r3} to extract hemoglobin and melanin pigments from a single skin color image is briefly reviewed in this section. Skin is a turbid media with multiple layers. Among various pigments, melanin and hemoglobin are dominantly contained in the epidermal and dermal layer, respectively. On the basis of the four assumptions made by Tsumura and Beer-Lambert law\cite{r5}, skin color can be modeled as a linear combination of the density of melanin and hemoglobin in the optical domain of three channels: $-\log (r_{l, m})$,$-\log (g_{l, m})$and$-\log (b_{l, m})$ (Fig.\ref{fig:optic}), where $r_{l, m}$, $g_{l, m}$ and $b_{l, m}$ denote the pixel values of the skin color image on the image coordinate $(l, m)$, respectively. It is seen that three densities of skin color are distributed on the two-dimensional plane spanned by pure color vectors melanin and hemoglobin. The color density vector $\boldsymbol{c}_{l, m}$ is denoted as
\begin{equation}
\boldsymbol{c}_{l, m}=\left[-\log \left(r_{l, m}\right),-\log \left(g_{l, m}\right),-\log \left(b_{l, m}\right)\right]^\top
\end{equation}
According to the skin color model shown in Fig.\ref{fig:optic}, the color density vector of skin can be expressed by
\begin{equation}
\label{eqn:color_density}
\boldsymbol{c}_{l, m}=\mathbf{C} \boldsymbol{q}_{l, m}
\end{equation}
where $\mathbf{C}=[\boldsymbol{c}(1), \boldsymbol{c}(2)], \boldsymbol{q}_{l, m}=\left[q_{l, m}(1), q_{l, m}(2)\right]^\top$. $\boldsymbol{c}(1)$ and $\boldsymbol{c}(2)$ denote density vector of melanin and hemoglobin, $q_{l, m}(1)$ and $q_{l, m}(2)$ are relative quantities of the pigments, $\boldsymbol{c}(3)$ is spatially stationary vector cause by other pigments and skin structure. From Eq.(\ref{eqn:color_density}), skin color can be considered as a mixed signal of two independent signals. PCA and ICA are then applied to extract the two-dimensional plane spanned by $\boldsymbol{c}(1)$ and $\boldsymbol{c}(2)$ and to estimate the quantity vector $\boldsymbol{q}_{l, m}$, respectively, for getting melanin distribution map and hemoglobin distribution map.

\subsection{Isomap {\it vs} PCA}
\label{ssec:isomap}
Classical scaling techniques, such as PCA and MDS, have proven to be successful in many applications, but it suffers from the fact that it mainly aims to retain pairwise Euclidean distances and thus fails to preserve the true structure of data lying on a complex nonlinear manifold (Fig.\ref{fig:isomap_vs_pca_a}). Isomap\cite{r4} is a technique that resolves this problem by attempting to preserve pairwise geodesic distances between datapoints. In Isomap, the geodesic distances between the datapoints $\mathbf{x}_{i}(i=1,2, \ldots, n)$ are computed by constructing a neighborhood graph $G$, in which every datapoint $\mathbf{x}_{i}$ is connected with its $k$ nearest neighbors $x_{i j}(j=1,2,\ldots,k)$ in the dataset $\mathbf{X}$. The shortest path between two points in the graph forms an estimate of the geodesic distance between these two points, and can easily be computed using Floyd's shortest-path algorithm\cite{r6}. The geodesic distances between all datapoints in $\mathbf{X}$ are computed, thereby forming a pairwise geodesic distance matrix. The low-dimensional representations $\mathbf{y}_{i}$ of the datapoints $\mathbf{x}_{i}$ in the low-dimensional space $\mathbf{Y}$ are computed by applying classical MDS on the resulting pairwise geodesic distance matrix. The computational complexity of Isomap is $O(n^{3})$ ($n$ is the number of datapoints), while PCA has a computational complexity of $O(D^{3})$ ($D$ is the dimensionality of datapoints). Thus, it will become a critical issue for Isomap to process high quality skin image (where $n \gg D, D \equiv 3$ ) in practical application.

\section{EXPERIMENT RESULTS}
\begin{figure}[t]
	\centering
	\subfigure[Synthetic 3D data ($50\times50$)]{
		\begin{minipage}[t]{0.438\linewidth}
			\includegraphics[width=1\linewidth]{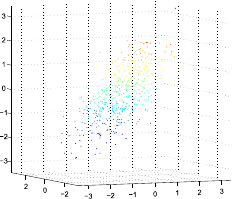} 
		\end{minipage}
		\label{fig:isomap_vs_pca_a}
	}
    	\subfigure[Data from Fig.\ref{fig:kim}(a) ($40\times32$)]{
    		\begin{minipage}[t]{0.438\linewidth}
   		 	\includegraphics[width=1\linewidth]{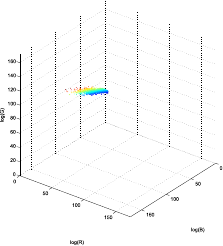}
    		\end{minipage}
		\label{fig:isomap_vs_pca_b}
    	}
	\\ 
	\subfigure[]{
		\begin{minipage}[t]{0.438\linewidth}
			\includegraphics[width=1\linewidth]{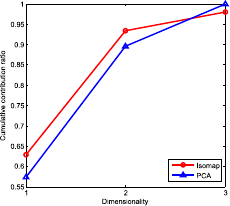} 
		\end{minipage}
		\label{fig:isomap_vs_pca_c}
	}
    	\subfigure[]{
    		\begin{minipage}[t]{0.438\linewidth}
		 	\includegraphics[width=1\linewidth]{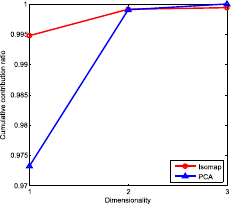}
    		\end{minipage}
		\label{fig:isomap_vs_pca_d}
    	}
	\caption{Comparison of dimensionality reduction performance between Isomap and PCA. (c), (d): cumulative contribution ratio on (a), (b).}
	\label{fig:isomap_vs_pca}
\end{figure}

In this section, we firstly Isomap and PCA on a set of synthetic data (Fig.\ref{fig:isomap_vs_pca_a}) and real skin data (Fig.\ref{fig:isomap_vs_pca_b}), respectively. Fig.\ref{fig:isomap_vs_pca_c} shows that Isomap outperforms PCA on dimensionality reduction of synthetic data with an accuracy of $93.4 \%$, while PCA preserves $89.5 \%$ of variance as measured in three dimensional nonlinear manifold. However, in application of pigment decomposition of small skin sample (Fig.\ref{fig:kim}(a)) in optical density domain, Isomap and PCA are both qualified in finding a two-dimension linear plane that best preserve the true structure of skin color data lying approximately on a linear manifold, with an accuracy of $99.9 \%$ (Fig.\ref{fig:isomap_vs_pca_d}).

\begin{figure}[t]
	\centering
	\subfigure[Regional facial skin]{
		\begin{minipage}[t]{0.398\linewidth}
			\includegraphics[width=1\linewidth]{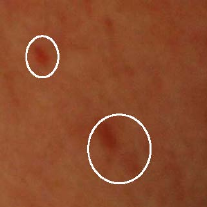} 
		\end{minipage}
		\label{fig:regional-facial}
	}
    	\subfigure[Entire facial skin]{
    		\begin{minipage}[t]{0.398\linewidth}
   		 	\includegraphics[width=1\linewidth]{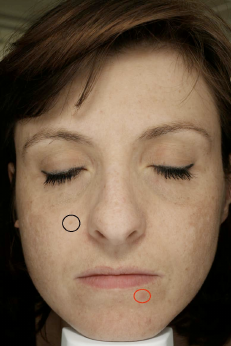}
    		\end{minipage}
		\label{fig:entire-facial}
    	}
	\caption{Experimental inputs: (a) white circle: pimple (b) black circle: freckle; red circle: pimple.}
	\label{fig:experiment_input}
\end{figure}

\begin{figure}[t]
	\centering
	\subfigure[]{
		\begin{minipage}[t]{0.338\linewidth}
			\includegraphics[width=1\linewidth]{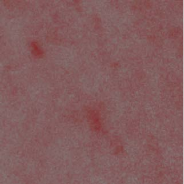} 
		\end{minipage}
		\label{fig:kim_hemo_region}
	}
    	\subfigure[]{
    		\begin{minipage}[t]{0.338\linewidth}
   		 	\includegraphics[width=1\linewidth]{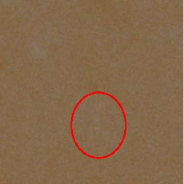}
    		\end{minipage}
		\label{fig:kim_melan_region}
    	}
	\\ 
	\subfigure[]{
		\begin{minipage}[t]{0.338\linewidth}
			\includegraphics[width=1\linewidth]{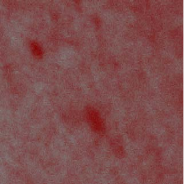} 
		\end{minipage}
		\label{fig:tsu_hemo_region}
	}
    	\subfigure[]{
    		\begin{minipage}[t]{0.338\linewidth}
		 	\includegraphics[width=1\linewidth]{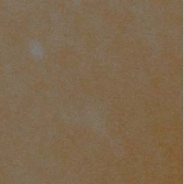}
    		\end{minipage}
		\label{fig:tsu_melan_region}
           }
  \\ 
	\subfigure[]{
		\begin{minipage}[t]{0.338\linewidth}
			\includegraphics[width=1\linewidth]{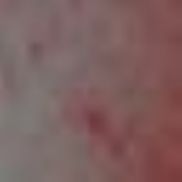} 
		\end{minipage}
		\label{fig:iso_hemo_region}
	}
    	\subfigure[]{
    		\begin{minipage}[t]{0.338\linewidth}
		 	\includegraphics[width=1\linewidth]{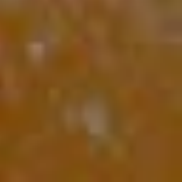}
    		\end{minipage}
		\label{fig:iso_melan_region}
    	}
	\caption{Pigment decomposition results of regional facial skin by Kim’s method: (a), (b); Tsumura’s method: (c), (d); Proposed method (Isomap+ICA): (e), (f). From left to right: hemoglobin map and melanin map. Red circle in (b) locates a mistake.}
	\label{fig:decompose_region}
\end{figure}

\begin{figure}[!htp]
	\centering
	\subfigure[]{
		\begin{minipage}[t]{0.472\linewidth}
			\includegraphics[width=1\linewidth]{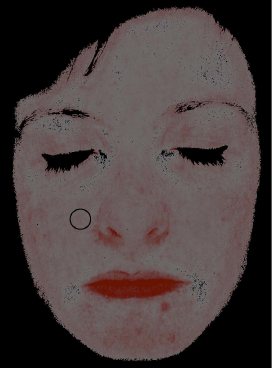} 
		\end{minipage}
		\label{fig:kim_hemo_entire}
	}
    	\subfigure[]{
    		\begin{minipage}[t]{0.472\linewidth}
   		 	\includegraphics[width=1\linewidth]{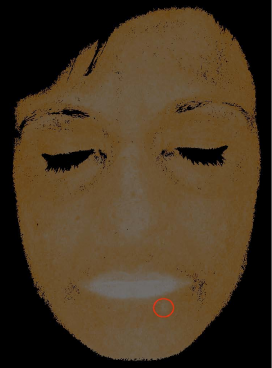}
    		\end{minipage}
		\label{fig:kim_melan_entire}
    	}
	\\ 
	\subfigure[]{
		\begin{minipage}[t]{0.472\linewidth}
			\includegraphics[width=1\linewidth]{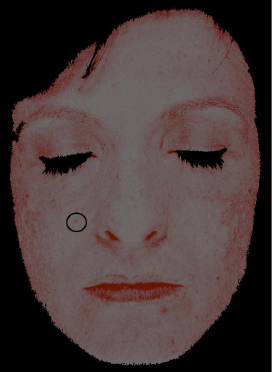} 
		\end{minipage}
		\label{fig:tsu_hemo_entire}
	}
    	\subfigure[]{
    		\begin{minipage}[t]{0.472\linewidth}
		 	\includegraphics[width=1\linewidth]{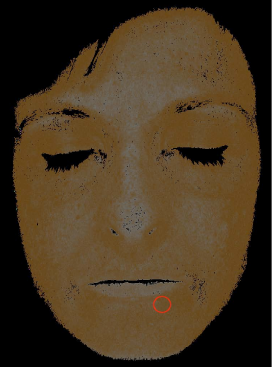}
    		\end{minipage}
		\label{fig:tsu_melan_entire}
    	}
	\caption{Pigment decomposition results of entire facial skin by Kim’s method: (a), (b) and Tsumura’s method: (c), (d). From left to right: hemoglobin map and melanin map.}
	\label{fig:decompose_entire}
\end{figure}

Skin pigment decomposition performances of the techniques mentioned in Section \ref{sec:decomp} are evaluated successively on two groups of experimental data (Fig.\ref{fig:experiment_input}): one is a $322\times322$ high-resolution image (for our proposed method, $32\times32$ low-resolution) of regional facial skin which contains only pimples; another is a $778\times1167$ high-resolution image (our method excluded) of entire facial skin which contains both freckles and pimples. Surface reflection that always produces highlight is removed by polarization filter in the two images. To our knowledge, the result of skin pigment decomposition can be evaluated by manually observing if freckles or pimples are successfully separated. Generally, freckles will only appear in melanin component while pimples appear only in hemoglobin component. 
Fig.\ref{fig:decompose_region} presents pigment decomposition results on regional facial skin (Fig.\ref{fig:regional-facial}). Here, Tsumura's method outperforms Kim's method as it extracts hemoglobin map much closer to the distribution of pimples, while the latter fails to extract the objective melanin map. Red circle mark (Fig.\ref{fig:kim_melan_region}) locates the false-separated part, which is supposed to contain no melanin at all. Though with lower resolution, our method still performs accurately on both hemoglobin and melanin parts (Fig.\ref{fig:iso_hemo_region}, \ref{fig:iso_melan_region}). 
Fig.\ref{fig:decompose_entire} shows the pigment decomposition results on entire facial skin (Fig.\ref{fig:entire-facial}). Here, Kim's method outperforms Tsumura's method in more accurate extraction of hemoglobin map and melanin map, while the latter makes incorrect separation on both pimple and freckle parts: black circled area in Fig.\ref{fig:tsu_hemo_entire} is supposed to have no hemoglobin at all and red circled area in Fig.\ref{fig:tsu_melan_entire} is supposed to have no melanin at all.
\section{DISCUSSION AND CONCLUSION}
\begin{figure}[!htp]
\centering
\begin{minipage}[t]{0.75\linewidth}
\centering
\centerline{\includegraphics[width=1.0\linewidth]{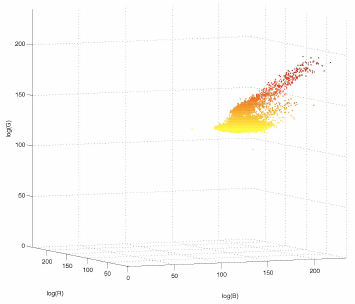}}
\end{minipage}
\caption{3D distribution of Fig.\ref{fig:entire-facial} (pure skin part).}
\label{fig:3d_entire}
\end{figure}

Essentially, both Kim's method and Tsumura's method adopt the same methodology to perform skin pigment decomposition, which contains three main steps: i) skin data modeling; ii) data dimensionality reduction; iii) independent feature extraction. At the first step, Kim's method adopts an intuitive color model, HSV color model, making the next two steps easier to be executed accurately without counting on any statistical techniques, such as PCA and ICA. Due to insufficient knowledge of optical skin properties, however, this method can give incorrect description of skin pigment quantity (Fig.\ref{fig:kim_melan_region}) and thus cannot be employed as an aide for medical diagnosis. Tsumura's method, on the other hand, is based on optical density model of human skin. Results obtained by this technique in Fig.\ref{fig:tsu_hemo_region} and \ref{fig:tsu_melan_region} agree quite well with the physiological knowledge. However, it performs incorrect skin pigment decomposition when applied to process the entire facial skin image (Fig.\ref{fig:tsu_hemo_entire}, \ref{fig:tsu_melan_entire}). Entire facial skin (Fig.\ref{fig:entire-facial}) data in optical density domain lies on a complex nonlinear manifold (Fig.\ref{fig:3d_entire}), which makes PCA fail in preserving the true structure of the data and thus results in incorrect skin pigment decomposition. Consequently, we introduce Isomap in combination with ICA, trying to solve this problem. In this paper, Isomap has proven theoretically and experimentally to be more accurate than PCA on complex nonlinear manifold. Unfortunately, due to its high computational complexity, our method is not applicable to large scale data sets. Future work aims to reduce the computational complexity of Isomap.

\bibliographystyle{IEEEbib}
\bibliography{refs}

\end{document}